\documentclass[letterpaper,10pt,conference]{ieeeconf}
\IEEEoverridecommandlockouts
\usepackage{amsmath,amssymb,graphicx,cite,booktabs}
\usepackage[table]{xcolor}
\usepackage{capt-of,flushend}
\usepackage[hidelinks]{hyperref}
\newcommand{\clip}{\operatorname{clip}}
\title{Escape Without a Recipe: Maneuver-Agnostic Mars Rover\\Recovery from Granular Entrapment}
\author{Meraj Hossain Promit$^{1}$, Chandak Chakma$^{1}$, Md Jubair Ahmed Sourov$^{1}$, and Sejuti Rahman$^{2}$%
\thanks{$^{1}$Department of Robotics \& Mechatronics Engineering, University of Dhaka, Bangladesh.}%
\thanks{$^{2}$Department of AI \& Robotics, New Uzbekistan University, Uzbekistan.}}
\IEEEaftertitletext{\begin{minipage}{\textwidth}
\centering
\includegraphics[width=\textwidth,trim=0 0 0 115.000bp,clip]{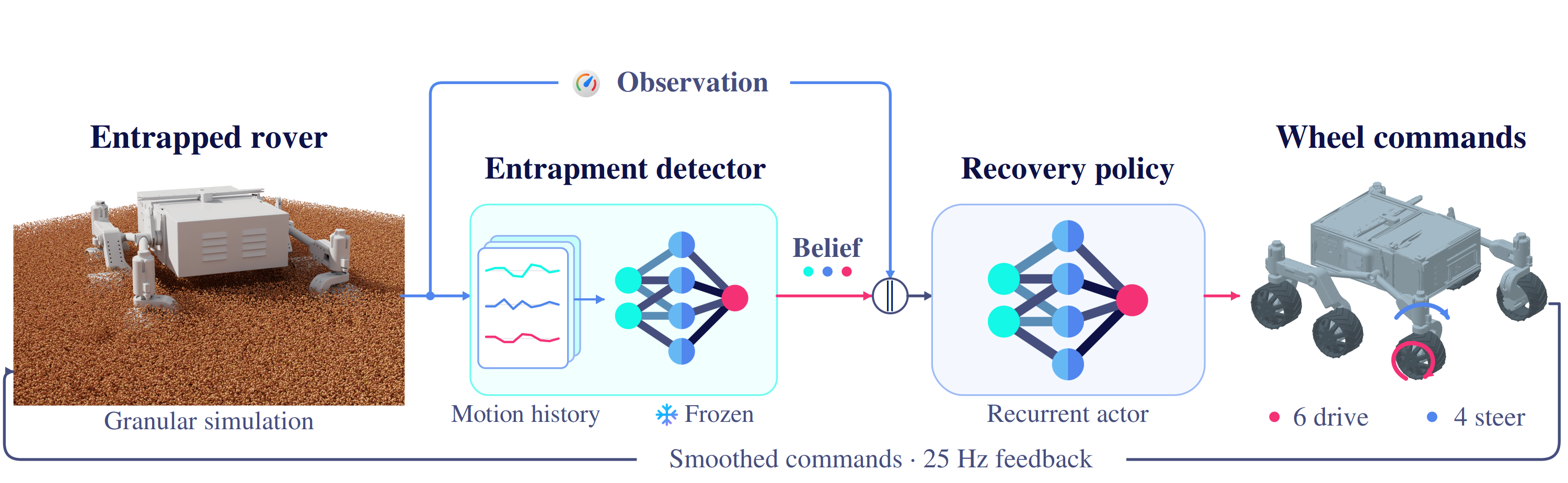}
\captionof{figure}{Recovery framework. A frozen entrapment detector supplies a three-state belief to a recurrent policy, which directly commands six drive and four steering joints at 25\,Hz. Left: simulated burial in the granular bed. Right: rover-mesh illustration; wheel contact uses smooth cylinders. Network nodes and traces are schematic. Network icon: Magnific/Flaticon.}
\label{fig:overview}
\end{minipage}\vspace{6pt}}
\hypersetup{pdftitle={Escape Without a Recipe: Maneuver-Agnostic Mars Rover Recovery from Granular Entrapment},pdfauthor={Meraj Hossain Promit, Chandak Chakma, Md Jubair Ahmed Sourov, Sejuti Rahman}}
\begin{document}
\maketitle
\begin{abstract}
Wheeled planetary rovers immobilised in loose regolith have no principled recovery controller: scripted maneuvers are terrain-specific, and naive driving deepens entrapment. We introduce a recurrent policy that learns to escape without a predefined maneuver. The framework rests on a verified trap: in a continuum material point method (MPM) simulator we identify a configuration at sinkage-to-radius ratio $z/r=2.0$ where the strongest of ten scripted controllers succeeds in only $7/32$ trials, yet the regolith retains traction and continues to shear, confirming granular entrapment rather than a numerical artifact. A frozen entrapment detector, trained on 81{,}771 windows spanning three materials and six sinkage levels, supplies a three-state belief to the policy observation. A separate simulator mobility latch gates training rewards. Across five training seeds, mean-action evaluation achieves $62.5\%$ escape (range $28.1$--$81.3\%$; 32 trials per seed) versus $7/32$ for the strongest scripted controller and $0/32$ for action noise. A numerical audit identifies a coupling-buffer fault that silently discarded $9.7\%$ of contact nodes at 16 parallel environments. Results establish recovery in the tested simulation regime; physical transfer remains unverified.
\end{abstract}

\section{Introduction}
Spirit's immobilization at Troy illustrates the difficulty of recovery after wheel burial~\cite{arvidson2010spirit}. Scripted approaches prescribe push-pull locomotion~\cite{creager2015pushpull} or material-remodeling gaits~\cite{shrivastava2020paddling}; optimization adapts their parameters~\cite{guo2024escape}. Reinforcement learning (RL) can instead discover commands directly. Prior learned extrication~\cite{yang2025extrication} and deformable-terrain locomotion~\cite{choi2023deformable,orsula2025sim2dust} motivate combining continuous recovery control with numerical checks of wheel--soil coupling.

Our framework (Figs.~\ref{fig:overview}--\ref{fig:loop}) learns recovery without demonstrations or prescribed maneuvers. We evaluate five training seeds against ten scripted controllers and audit numerical coupling.

\begin{figure*}[t]
\centering
\includegraphics[width=\textwidth]{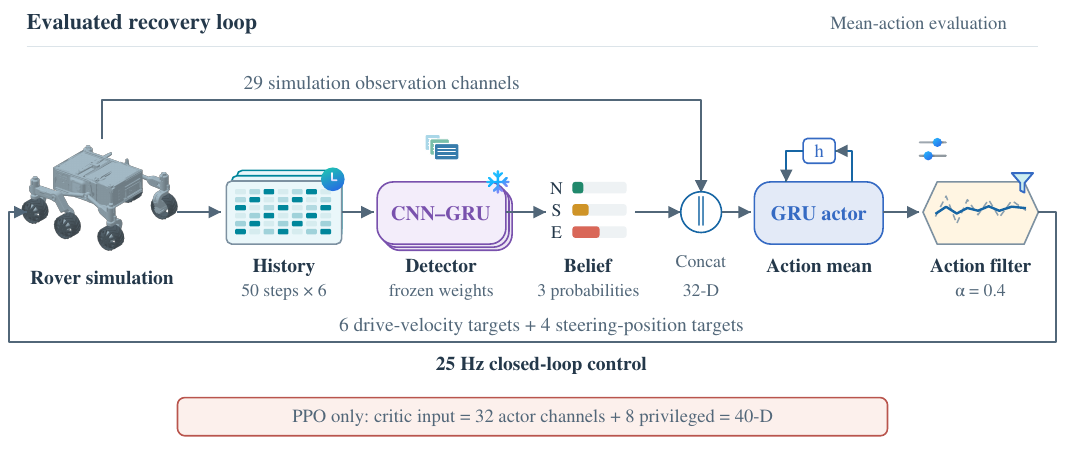}
\caption{Evaluated recovery loop. A frozen convolutional neural network (CNN)--gated recurrent unit (GRU) maps motion history to normal, sinking, and entrapped probabilities. These augment 29 simulation channels; the normalized 32-D actor input produces ten commands, filtered with $\alpha=0.4$ at 25\,Hz. The training-only critic receives eight additional privileged features. Normalization and scaling are omitted; traces and probabilities are schematic.}
\label{fig:loop}
\end{figure*}

\begin{figure*}[t]
\centering
\includegraphics[width=0.94\textwidth]{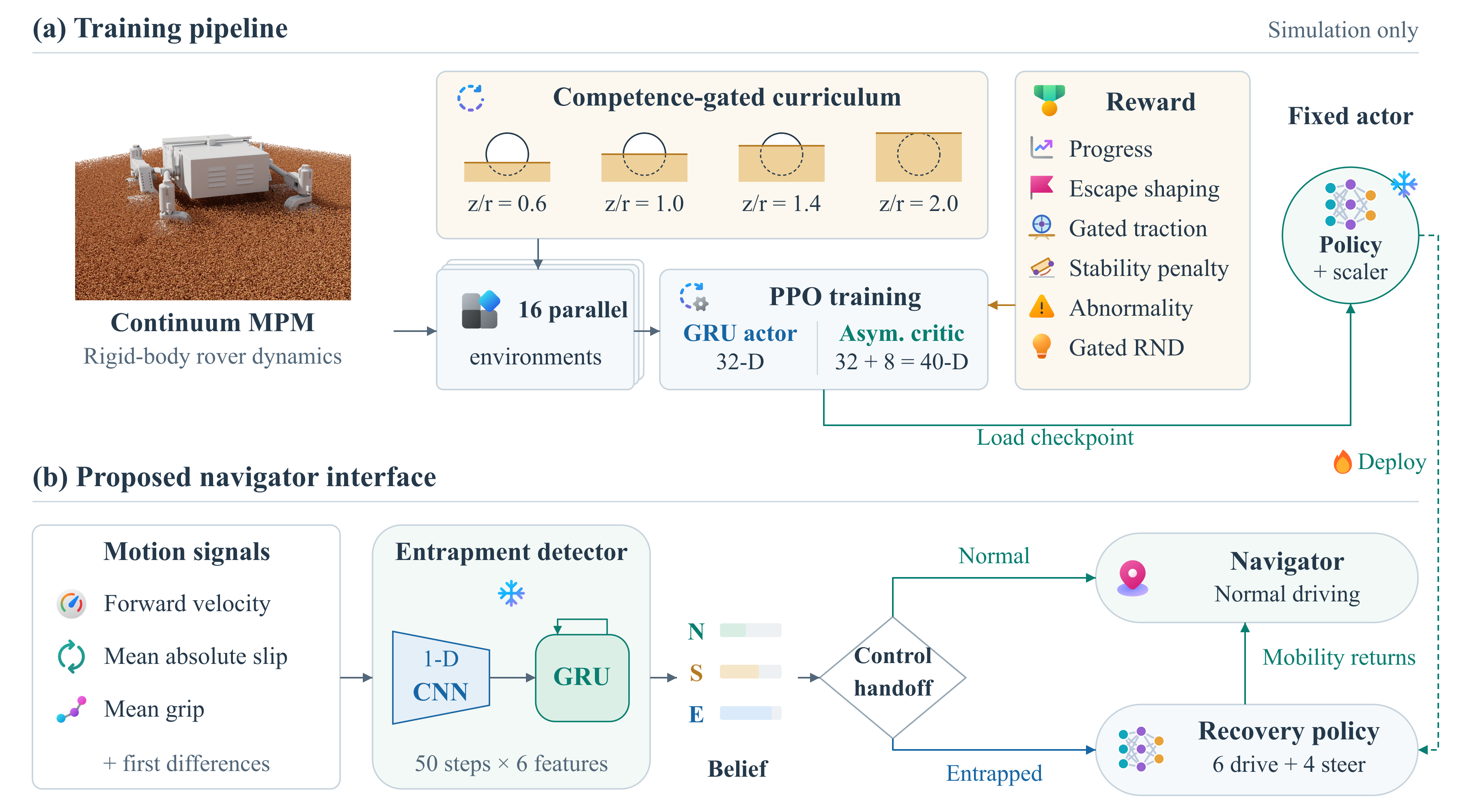}
\setcounter{figure}{2}
\caption{Training pipeline and proposed navigator interface. Top: a competence-gated burial curriculum supplies 16 parallel MPM environments; proximal policy optimization (PPO) trains the actor and critic with random network distillation (RND). Bottom: proposed navigation/recovery handoff, not evaluated. Experiments assess the recovery policy alone.}
\label{fig:training}
\end{figure*}
\setcounter{figure}{3}

\section{Simulation and Recovery Definition}
\subsection{Rover and Granular Terrain}
We simulate a $33.4$\,kg rover with six driven wheels and four corner-steering joints. Wheel radius is $r=0.0994$\,m; collision geometry uses smooth cylinders. The rover mesh follows the RLRoverLAB model~\cite{rlroverlab}. A $3.5\times3.5\times0.6$\,m granular bed uses continuum material points with an implicit MPM solver, coupled two-way with rigid-body dynamics in the NEWTON physics engine~\cite{sulsky1994particle}.

Gravity is $3.72$\,m\,s$^{-2}$. A jittered lattice at $1700$\,kg\,m$^{-3}$ uses $2.5$\,cm particle spacing and a $5$\,cm grid: $7.95$ million particles across 16 environments. Frictional rheology uses $500$\,Pa yield stress, $\mu_w=0.3$, and internal friction sampled over $0.58$--$0.75$ per reset. The solve uses at most 100 iterations at tolerance $10^{-7}$.

At $z/r=2$, burial is approximately $0.20$\,m. Training varies initial burial; reporting fixes depth. The $33.4$\,kg platform is not a scale model of the $174$\,kg Spirit.

\subsection{Slip and Sensing Assumptions}
For wheel $i$ with ground-relative velocity $v_i$ and angular velocity $\omega_i$, the longitudinal slip and grip are
\begin{equation}
\begin{aligned}
\sigma_i&=\clip\!\left(\frac{r\omega_i-v_i}{\max(|r\omega_i|,|v_i|,0.05)},-1,1\right),\\
g&=1-\frac{1}{6}\sum_{i=1}^6|\sigma_i|.
\end{aligned}
\label{eq:slip}
\end{equation}
The velocity floor is in m\,s$^{-1}$. Wheel velocities and actor displacement use simulator state; hardware requires independent velocity estimation. Hardware sensing is untested. Grip is a slip statistic, not measured force.

\subsection{Operational Mobility Criterion}
A trial succeeds within a 40\,s horizon if either (i) mean forward speed exceeds $0.15$\,m\,s$^{-1}$ and mean grip exceeds $0.35$ over a 50-step (2\,s) window, or (ii) net displacement along the escape heading exceeds $3$\,m and window-mean grip exceeds $0.05$. Both paths require the window to exclude initial burial grace and its maximum absolute forward speed to remain below $1.5$\,m\,s$^{-1}$, rejecting transient numerical ejections.

These empirical thresholds reject brief transients. Threshold sensitivity is untested.

\section{Learning Maneuver-Agnostic Recovery}
\subsection{Detector and Recurrent Actor}
The CNN--GRU detector uses 50 steps of velocity, absolute slip, grip, and their differences. A split of 81,771 windows across three materials and six depths gives 68,092 training and 13,679 validation windows. Macro-F1 is $0.966$; material overlap limits this checkpoint-selection score.

The actor receives 29 channels (wheel speed, slip, steering, inertial signals, torque changes, flags, projected displacement) plus three beliefs; its asymmetric critic~\cite{pinto2018asymmetric} receives eight more. Evaluation retains normalization and filtering; recurrent states reset per episode.

\subsection{Progress-Gated Reward}
The reward combines signed progress, escape shaping~\cite{ng1999shaping} (including a one-time $+20$ mobility bonus), gated traction, tilt/abnormality penalties, and competence-scaled random network distillation (RND)~\cite{burda2019rnd}. The simulator mobility latch, not detector belief, gates traction and novelty. Fling guards suppress positive rewards. For heading velocity $v_t$, $\Delta t=0.04$\,s, and gate $f_t$, the objective is
\begin{equation}
R_t=8v_t\Delta t+r_t^{\rm esc}+90f_t r_t^{\rm trac}-p_t
+\beta_t f_t r_t^{\rm RND}.
\label{eq:reward}
\end{equation}
For heading displacement $d_t$ and its previous maximum $d^*_{t-1}$, the traction term is
\begin{equation}
 r_t^{\rm trac}=g_t e_t q_t\Delta t\,
 \clip\!\left(\frac{(d_t-d^*_{t-1})_+/\Delta t-v_0}{v_{\rm ref}},0,1\right),
\label{eq:traction}
\end{equation}
where $(x)_+=\max(x,0)$, $v_0=0.02$\,m\,s$^{-1}$, $v_{\rm ref}$ is full-throttle rim speed, $q_t$ is the non-fling indicator, and $e_t=\clip((\overline{|r\omega|}-0.05)/0.10,0,1)$ gates drive engagement. The maximum starts at zero and updates each step. Repeating previously reached excursions earns no traction reward.

\begin{figure}[t]
\hrule\vspace{2pt}
{\small\textbf{Algorithm 1} Recurrent recovery training}\par
\vspace{2pt}\hrule\vspace{3pt}
{\footnotesize\renewcommand{\arraystretch}{0.92}
\begin{tabular}{@{}r@{\quad}p{0.89\columnwidth}@{}}
\multicolumn{2}{@{}l@{}}{\textbf{Input:} frozen detector $D$, actor $\pi_\theta$, critic $V_\phi$;}\\
\multicolumn{2}{@{}l@{}}{16 environments, 100k control steps/environment.}\\
1 & Initialize observations, recurrent and filter states.\\
2 & \textbf{while} the control-step budget remains \textbf{do}\\
3 & \quad Collect 256 steps/env: $b_t{=}D(\text{sensor history})$; normalize $o_t{=}[x_t;b_t]\in\mathbb{R}^{32}$.\\
4 & \qquad Sample $a_t{\sim}\mathcal{N}(\mu_t,\operatorname{diag}\sigma_a^2)$; filter $u_t{=}0.4a_t{+}0.6u_{t-1}$ at 25\,Hz.\\
5 & \qquad Receive reward (\ref{eq:reward}), next obs, done flags; env updates latch.\\
6 & \qquad Clear recurrent/filter states for reset episodes.\\
7 & \quad Compute GAE; bootstrap 40\,s truncations. Update $\pi_\theta,V_\phi$ with PPO.\\
8 & \quad Update the curriculum on environment resets.\\
9 & \textbf{end while}; \textbf{return} $\pi_\theta$ and normalization.\\
\multicolumn{2}{@{}l@{}}{\textbf{Evaluate:} filtered mean action; retain $D$ and normalization;}\\
\multicolumn{2}{@{}l@{}}{discard critic and curriculum.}\\
\end{tabular}}\par\vspace{3pt}\hrule
\end{figure}

\subsection{Training Budget}
Algorithm~1 and Fig.~\ref{fig:training} summarize PPO training~\cite{schulman2017ppo}: discount $0.99$, generalized-advantage parameter $0.95$, clipping $0.2$, initial learning rate $3\times10^{-4}$, three epochs, eight minibatches, 64-step backpropagation through time (BPTT), entropy weight $0.003$, and initial action log standard deviation $-0.5$.

Each seed receives 100k steps/environment ($1.6$ million transitions). Each uninterrupted run required $29$--$29.2$ hours on an RTX 4090, approximately 145 hours for five seeds. The curriculum advances through $z/r\in\{0.6,0.8,\ldots,2.0\}$ at $60\%$ escape, following reverse curriculum learning~\cite{florensa2017reverse}.

\section{Experimental Results}
\begin{figure*}[t]
\centering
\includegraphics[width=0.98\textwidth]{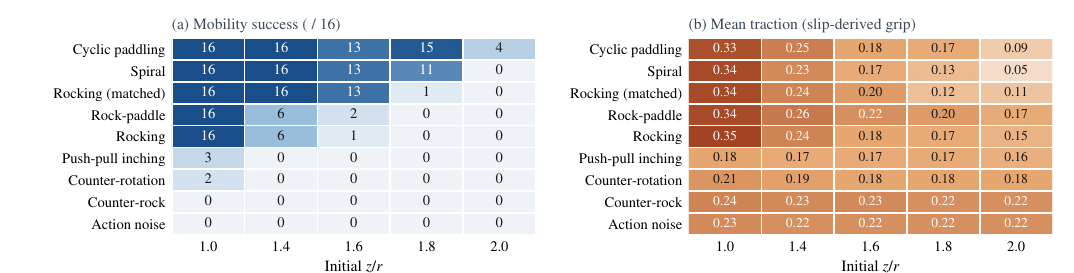}
\caption{Depth screening. \emph{(a)} Recoveries/16 for eight scripted controllers and action noise. \emph{(b)} Slip-derived grip. At $z/r=2.0$ all arms except paddling score zero, yet grip stays nonzero ($0.05$--$0.22$): the regolith still carries traction, confirming granular rather than numerical entrapment.}
\label{fig:depth}
\end{figure*}

\subsection{Protocol and Scripted Comparisons}
Checkpoints at 60/70/80/90/100k steps are selected using mean-action recovery at $z/r=2.0$ on evaluation seed 0 (16 trials). Reporting uses evaluation seed 1 (32 trials per depth and mode). Five uninterrupted training seeds reached the 100k budget; a sixth interrupted run had a different effective budget.

The ten scripted controllers are constant drive, push-pull inching, cyclic-sweep paddling, rocking, rock-paddle, spiral, crab walk, crab shear, squirm, and counter-rotation. Native and duty-cycle variants account for differing command intensities. Figure~\ref{fig:depth} shows depth screening: at $z/r=2.0$ all arms except paddling fail, yet mean grip stays nonzero ($0.05$--$0.22$), confirming the regolith remains mechanically active. A separate corrected-coupling constant-drive screen decreases from $16/16$ escapes at $z/r=1.0$ to $0/16$ at $2.0$.

\begin{figure}[t]
\centering
\includegraphics[width=\columnwidth]{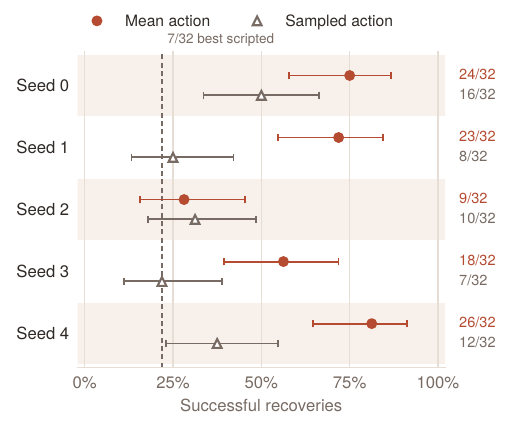}
\setcounter{figure}{4}
\caption{Training-seed and action-mode variability at $z/r=2.0$. Points show recoveries/32, with pointwise 95\% Wilson intervals. Dashed lines mark the best scripted rate ($7/32$). Rows share training seeds, not paired trajectories.}
\label{fig:performance}
\end{figure}

Fig.~\ref{fig:performance} shows a mean escape rate of $62.5\%$ across five seeds, with a $28.1$--$81.3\%$ range. At $z/r=1.8$, the same mean-action checkpoints average $80.0\%$. The spread limits training-reliability claims. Sampled-action evaluation averages $33.1\%$ at the trap, indicating sensitivity to deployment action mode.

Zero-mean Gaussian action noise yields $0/32$ escapes and $-0.004$\,m mean displacement despite higher mean grip ($0.22$) than the policy seeds ($0.11$--$0.20$). Grip alone is insufficient. Noise variance was not matched to every checkpoint.

\subsection{Detector-Belief Ablation}
Table~\ref{tab:ablation} compares seed-0 training without belief or with raw $P_{\rm ent}$ replacing the reward gate. Selected checkpoints are 80k (full), 70k (no belief), and 80k (raw gate); the raw-probability run resumed after interruptions.

\begin{table}[t]
\caption{Detector-belief ablation. Recoveries/32 per cell in a separate single-training-seed comparison.}
\label{tab:ablation}
\centering\small
\setlength{\tabcolsep}{3pt}\renewcommand{\arraystretch}{1.15}
\begin{tabular*}{\columnwidth}{@{\extracolsep{\fill}}lrrrr@{}}
\toprule
&\multicolumn{2}{c}{$z/r=1.8$}&\multicolumn{2}{c}{$z/r=2.0$}\\
\cmidrule(lr){2-3}\cmidrule(l){4-5}
Training variant & Mean & Sampled & Mean & Sampled\\
\midrule
\rowcolor[HTML]{EBF5F7}
Full policy & 32/32 & 27/32 & 23/32 & 14/32\\
Without belief & 31/32 & 29/32 & 19/32 & 11/32\\
Raw-$P_{\rm ent}$ gate & 16/32 & 11/32 & 12/32 & 5/32\\
\bottomrule
\end{tabular*}
\end{table}

The full policy performs best at $z/r=2.0$; removing belief improves sampled-action recovery at $z/r=1.8$. This single-seed comparison does not establish a consistent benefit.

\section{Numerical Integrity and Limitations}
A coupling buffer capped transfers at $2^{20}$ nodes, discarding $9.7\%$ of contact nodes at 16 environments without an error. This does not imply the same fraction of lost force. We check environment-count consistency, coupling conservation (truncation raises an error), and solver-array parameter read-back. A pre-fix material sweep does not establish corrected-coupling cross-material robustness.

Numerical consistency does not validate material behavior, wheel contact, or grid convergence. Finite bed depth, smooth wheels, imposed burial, privileged sensing, and mass scale constrain transfer. Hardware validation, cross-material tests, sensing-error robustness, and threshold sensitivity remain future work.

\section{Conclusion}
We presented a recurrent policy that recovers a rover from deep granular entrapment without prescribed escape maneuvers. A frozen detector augments the actor's observation with a three-state belief, while a simulator mobility latch gates training rewards. Across five seeds at a verified trap ($z/r=2.0$), mean-action evaluation achieves $62.5\%$ escape (range $28.1$--$81.3\%$), versus $21.9\%$ for the strongest scripted controller and $0/32$ for action noise. The depth screen confirms the regolith retains traction when scripted arms fail, distinguishing granular entrapment from a numerical artifact. A single-seed ablation shows removing detector belief degrades escape ($19/32$ vs.\ $23/32$), while raw-probability gating collapses it ($12/32$).

The measurement-integrity audit proved the principal obstacle: a coupling buffer silently discarded $9.7\%$ of contact nodes at $16$ environments. The three invariants we report expose this fault class in minutes. Training variability is significant, the trap uses a single material cell, and the $33.4$\,kg platform is not a scale model of Spirit. Hardware validation, cross-material generalization, and resolution convergence remain future work; physical transfer is not claimed.


{\scriptsize
\begin{thebibliography}{99}
\setlength{\itemsep}{-2pt}\setlength{\parskip}{0pt}
\bibitem{arvidson2010spirit} R.~E. Arvidson \emph{et al.}, ``Spirit Mars rover mission: Overview and selected results from the northern Home Plate Winter Haven to the side of Scamander crater,'' \emph{J. Geophys. Res. Planets}, vol.~115, 2010.
\bibitem{creager2015pushpull} C.~Creager, K.~Johnson, M.~Plant, S.~Moreland, and K.~Skonieczny, ``Push-pull locomotion for vehicle extrication,'' \emph{J. Terramech.}, vol.~57, pp.~71--80, 2015.
\bibitem{shrivastava2020paddling} S.~Shrivastava \emph{et al.}, ``Material remodeling and unconventional gaits facilitate locomotion of a robophysical rover over granular terrain,'' \emph{Sci. Robot.}, vol.~5, no.~42, 2020.
\bibitem{guo2024escape} J.~Guo \emph{et al.}, ``An online optimization escape entrapment strategy for planetary rovers based on Bayesian optimization,'' \emph{J. Field Robot.}, vol.~41, no.~8, pp.~2518--2529, 2024.
\bibitem{yang2025extrication} W.~Yang, Y.~Xing, and H.~Zhang, ``A method for autonomous intelligent extrication control of Mars rover based on deep reinforcement learning,'' in \emph{Advances in Computational Vision and Robotics}, Springer, 2025, pp.~122--132.
\bibitem{choi2023deformable} S.~Choi, G.~Ji, J.~Park, H.~Kim, J.~Mun, J.~H. Lee,
  and J.~Hwangbo, ``Learning quadrupedal locomotion on deformable terrain,''
  \emph{Sci. Robot.}, vol.~8, no.~74, eade2256, 2023.
\bibitem{orsula2025sim2dust} A.~Orsula, M.~Geist, M.~Olivares-Mendez, and
  C.~Martinez, ``Sim2Dust: Mastering dynamic waypoint tracking on granular
  media,'' in \emph{Int. Conf. Space Robotics (iSpaRo)}, 2025.
\bibitem{rlroverlab} ``RLRoverLAB: An advanced reinforcement learning suite for planetary rover simulation and training,'' \emph{IEEE}, 2025.
\bibitem{sulsky1994particle} D.~Sulsky, Z.~Chen, and H.~L. Schreyer, ``A particle method for history-dependent materials,'' \emph{Comput. Methods Appl. Mech. Eng.}, vol.~118, pp.~179--196, 1994.
\bibitem{pinto2018asymmetric} L.~Pinto, M.~Andrychowicz, P.~Welinder,
  W.~Zaremba, and P.~Abbeel, ``Asymmetric actor critic for image-based robot
  learning,'' in \emph{Robotics: Science and Systems}, 2018.
\bibitem{ng1999shaping} A.~Y. Ng, D.~Harada, and S.~Russell, ``Policy invariance under reward transformations: Theory and application to reward shaping,'' in \emph{Int. Conf. Machine Learning}, 1999.
\bibitem{burda2019rnd} Y.~Burda, H.~Edwards, A.~Storkey, and O.~Klimov, ``Exploration by random network distillation,'' in \emph{Int. Conf. Learning Representations}, 2019.
\bibitem{schulman2017ppo} J.~Schulman, F.~Wolski, P.~Dhariwal, A.~Radford, and O.~Klimov, ``Proximal policy optimization algorithms,'' arXiv:1707.06347, 2017.
\bibitem{florensa2017reverse} C.~Florensa, D.~Held, M.~Wulfmeier,
M.~Zhang, and P.~Abbeel, ``Reverse curriculum generation for reinforcement
learning,'' in \emph{Proc. Conf. Robot Learning}, vol.~78, 2017, pp.~482--495.
\end{thebibliography}}
\end{document}